# Learning Multimodal Fixed-Point Weights using Gradient Descent

Lukas Enderich[1], Fabian Timm[1], Lars Rosenbaum[1] and Wolfram Burgard[2]

1- Robert Bosch GmbH - Automated Driving Research

2- University of Freiburg - Autonomous Intelligent Systems

**Abstract**. Due to their high computational complexity, deep neural networks are still limited to powerful processing units. To promote a reduced model complexity by dint of low-bit fixed-point quantization, we propose a gradient-based optimization strategy to generate a symmetric mixture of Gaussian modes (**SGM**) where each mode belongs to a particular quantization stage. We achieve 2-bit state-of-the-art performance and illustrate the model's ability for self-dependent weight adaptation during training.

## 1 Introduction

Deep architectures with stacked non-linearities have highly pushed the progress of machine learning in the recent past. In this context, deep neural networks (DNNs) have set benchmarks in various fields of research, such as computer vision, speech recognition and image classification [1]. However, depending on millions of parameters and billions of high-precision computations [2, 3], DNNs require powerful and expensive processing units.

To address the problem of complexity, strict weight compression techniques using two bit [4, 5, 6] or even one bit [7, 6] have been developed. This reduces storage costs significantly and replaces floating-point multiplications by additions. However, most quantization functions keep high-precision scaling coefficients [4, 5, 6, 8], which eliminates the posibility of pure fixed-point arithmetic on dedicated hardware. In addition, incremental retraining steps, like weight partitioning [6, 8], complicate the training procedure and involve additional hyperparameters.

We propose a gradient-based optimization strategy to train DNNs with multimodal weight distributions that enable accurate post-quantization. While using pure fixed-point quantization, self-reliant weight adaptation makes retraining unnecessary. The loss function is easy to implement and achieves state-of-the-art performance using ternary weights (6.19% on CIFAR-10).

## 2 Related Work

*Hard quantization* methods use discrete weights in at least parts of the network flow to minimize the loss pertubation that is caused by quantization. In 2015, Courbariaux *et al.* combined weight binarization ($w_i^{\text{bin}} \in \{-1, 1\}$) and backpropagation [7]. While quantizing during forward and backward pass, high-precision weights are kept and updated subsequently. Doing so, the training with gradient descent converges. Li *et al.* increased the model capacity by combining ternary-valued weights and a scaling coefficient $\alpha$, $w_i^{\text{ter}} \in \{-\alpha, 0, \alpha\}$ [4]. To find an optimal value for $\alpha$, the euclidean distance between $\mathbf{W}$ and $\mathbf{W}^{\text{ter}}$ is approximated and solved with a threshold based ternary function. A straight forward extension was introduced by Zhou *et al.* by using two independent scaling factors $w_i^{\text{ter}} \in \{-\alpha^n, 0, \alpha^p\}$ and gradients to update both the continuous weights $w$ and the scaling coefficients $\alpha^n$ and $\alpha^p$ [5].

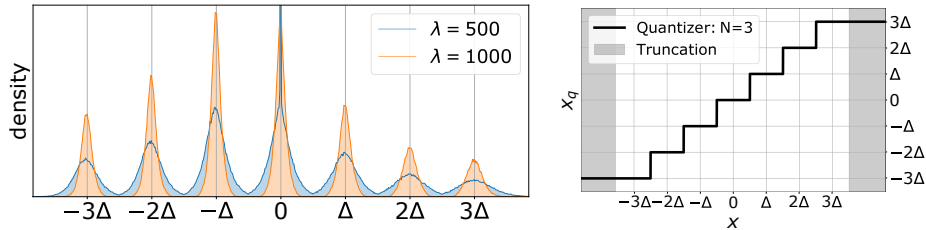

Fig. 1: Left(a): Weight distributions consisting of symmetric Gaussian modes. The corresponding variances result from different values of the regularization parameter $\lambda$. Right(b): Dedicated $N = 3$bit quantizer with uniform stepsize $\Delta$.

*Soft quantization* describes a training with real-valued weights, promoting posterior distributions that are well-suited for post-quantization. In this context, Bayesian methods have been used for model compression with the objective of sparse posterior distributions [9, 10]. Recently, Achterhold *et al.* introduced a quantizing prior to train DNNs with multimodal weight distributions that can be quantized with little loss in accuracy using ternary-valued weights [11]. Furthermore, Zhou *et al.* launched a loss-error-aware quantization method for determinsistic low bit DNNs [6]. Their key feature is to partition the layer weights in several steps, quantizing only one part while retraining the rest.

In order to omit high-precision scaling coefficients and retraining, we propose a single shot gradient based optimization strategy to train network weights that fit pure fixed-point arithmetic. We show that our model is a composition of symmetric Gaussian modes where the amount of modes and their particular variance can be regulated as required, see Figure 1(a).

Very recent work on arXiv [12] also describes the regularization of quantized weights. However, we introduce a reasonable layer-dependent gradient scale and demonstrate its benefit on weight adaptation.

## 3 Multimodal Fixed-point Weights

Initially, we illustrate the connection between symmetric quantization functions and fixed-point numbers and propose a suitable optimization strategy afterwards. We analyze the gradient behavior, explain differences to related work [6] and highlight the relation to $\mathcal{L}_2$ regularization.

### 3.1 Bit Shift Quantization

The representation of an $N$-bit fixed-point number is $(-1)^s \times m \times 2^{-f}$, where $(-1)^s \times m$ is the signed integer mantissa, $f \in \mathbb{Z}$ the position of the decimal point and $2^{-f}$ the global scaling factor [13]. This means that a scaling by a power of two is equivalent to moving the decimal point respectively. To exploit the *bit shift*, we use the following symmetric and uniform quantization structure

$$x_{\mathrm{q}} = Q_N(x; \Delta) = \underbrace{\mathrm{Clip}\left(\left\lfloor \frac{x}{\Delta} \right\rceil, -2^{N-1} + 1, \, 2^{N-1} - 1\right)}_{N\text{-bit signed integer}} \Delta \tag{1}$$

where $\lfloor \cdot \rceil$ describes the rounding to the closest integer, $\Delta$ is the uniform step size, Clip(x, min, max) truncates all values to the domain [min, max] and $x_q$ represents the quantized value. As an example, $Q_3$ is visualized in Figure 1(b). We see that Equation 1 corresponds to the fixed-point representation if and only if the step size is a power of two, hence $\Delta = 2^{-f}$, $f \in \mathbb{Z}$. In that case, $x_{\mathrm{q}}$ can be stored as $N$-bit fixed point number without loss in accuracy.

### 3.2 Symmetric & Multimodal Gaussian Prior

DNNs are commonly trained by minimizing a loss function $\mathcal{L}$ via gradient descent and backpropagation. To influence the resulting weight distribution, $\mathcal{L}$ can be amplified by a regularization $\mathcal{L}_{\mathrm{R}}$. In that case, one updates a single weight value $w_i$ in the negative direction of both gradients, scaled by the learning rate $\eta$

$$w_i \leftarrow w_i - \eta \left( \frac{\partial \mathcal{L}}{\partial w_i} + \frac{\partial \mathcal{L}_{\mathrm{R}}}{\partial w_i} \right). \tag{2}$$

$\mathcal{L}_2$ regularization is frequently used to give a Gaussian prior on the network weights and to prevent high filter energies [14]. In order to combine multiple Gaussian modes and the *bit shift* quantizer, we propose

$$\mathcal{L}_{\mathrm{R}} = \sum_{l=1}^{L} \sum_{i=1}^{M^{(l)}} \frac{\lambda}{M^{(l)}} \mathcal{L}_2\left(w_i^{(l)} - Q_N(w_i^{(l)}; \Delta^{(l)})\right) = \sum_{l=1}^{L} \sum_{i=1}^{M^{(l)}} \frac{\lambda}{2M^{(l)}} \left(w_i^{(l)} - w_{i,q}^{(l)}\right)^2$$

where $L$ is the number of layers, $M^{(l)}$ the number of weights in layer $l$, $\Delta^{(l)} = 2^{-f^{(l)}}$ the step size in layer $l$, and $\lambda$ the regularization parameter. The quantized version of $w_i^{(l)}$ is called $w_{i,q}^{(l)}$. The gradient respective $w_i^{(l)}$ is

$$\frac{\partial \mathcal{L}_{\mathrm{R}}}{\partial w_i^{(l)}} = \frac{\lambda}{M^{(l)}} \left(w_i^{(l)} - w_{i,q}^{(l)}\right) \left(1 - \frac{\partial w_{i,q}^{(l)}}{\partial w_i^{(l)}}\right) = \begin{cases} \pm\infty & \text{if } w_i^{(l)} = (k + \frac{1}{2})\Delta^{(l)} \\ \lambda/M^{(l)}(w_i^{(l)} - w_{i,q}^{(l)}) & \text{else,} \end{cases} \tag{3}$$

where $\partial w_{i,q}^{(l)}/\partial w_i^{(l)}$ is zero except between neighboring stages (see Figure 1(b)). The singularity can be neglected due to real-valued layer weights. This property is beneficial since $Q_N$ does not have to be smooth and can, however, ensure that a fixed-point representation is pursued.

The quantizing gradient used in [6] is a sign function scaled by a constant parameter. This is intuitively unqualified for convergence since each weight is updated with the same step size and needs to be fixed manually at the quantization stage. In contrast, our gradient supports self-supervised convergence. Furthermore, [6] applies incremental retraining steps in both dimensions (network width and depth) to repair the resulting accuracy degradation. We use the layer wise mean and scale the gradient with $\lambda/M^{(l)}$ to rate the quantization error of individual layers equally. Doing so, layers with many weights, which are more robust against quantization noise, can be provided with greater flexibility to compensate the overall compression loss. Applying layer-dependent gradient scales is a novel feature and proves to be necessary for self-reliant weight adaptation.

## 4 Experiments and Analysis

We compare our Symmetric Gaussian Mixture (**SGM**) with state-of-the-art weight compression results from BinnaryConnect (BC [7]), Ternary Weight Networks (TWN [4]), Trained Ternary Quantization (TTQ [5]), Explicit Loss-Error-Aware Quantization (ELQ [6]) and Variational Network Quantization (VNQ [11]). We use the same pre-processing steps and equal or smaller network architectures as in [4, 7, 5, 6, 11]. The regularization parameter $\lambda$ is chosen empirically using the training set. A suitable value for $\Delta^{(l)}$ can be determined by minimizing $\mathcal{L}_R$ either in the early stages of training or based on pretrained network weights. FP indicates if the quantizer uses fixed-point arithmetic.

### 4.1 LeNet-5 on MNIST

MNIST is a handwritten-digits classification task with $28\times28$ gray scale images divided into 50,000 training and 10,000 test samples [15]. We use LeNet-5 [16, 11] and initialize our network with pretrained weights (0.7% test error). We use SGD optimization for 80 epochs in combination with a batch size of 64.

|  | Bits | Error | FP |
|---|---|---|---|
| BC | 1 | 1.29% | ✓ |
| TWN | 2 | 0.65% | ✗ |
| VNQ | 2 | 0.73% | ✗ |
| **SGM** | 2 | **0.63%** | ✓ |

Table 1: Test error on MNIST.

The learning rate is linearly decreased from 0.01 to 0.001, $\lambda$ increases from 0 to 1000. Results are shown in Table 1. Our SGM approach yields state-of-the-art performance, while using pure fixed-point quantization.

### 4.2 DenseNet & VGG-7 on CIFAR-10

CIFAR-10 is an image classification benchmark data set, consisting of $32\times32$ RGB pictures with 50,000 training and 10,000 test samples [17]. For reliable comparisons, we use two different network architectures. First, a conventional CNN with seven layers called VGG-7 [4], second DenseNet (L = 76, k = 12) which has an optimized architecture with comparatively less parameters [18]. Both networks are initialized with pretrained weights (VGG-7: Error 6.12%, DenseNet: Error 5.68%) and optimized for 250 epochs using SGD optimization and a batch size of 64. The learning rate is initialized with 0.02 and linearly decreased to 0.002 while $\lambda$ rises fro 0 to 2000. Results are shown in Table 2.

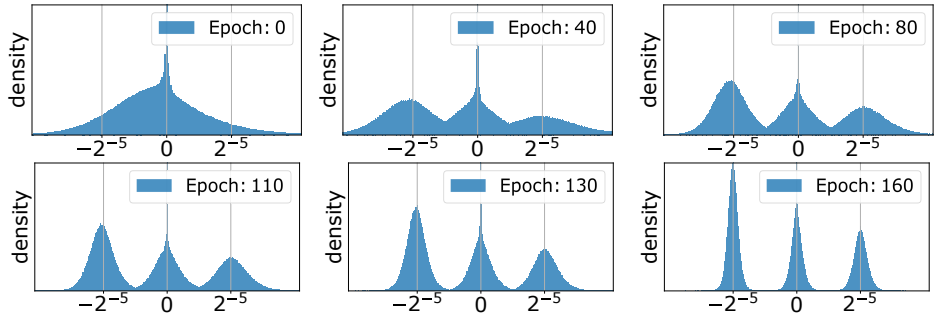

Fig. 2: Weight distribution of $Conv_5$ (VGG-7, CIFAR-10) for several epochs.

**VGG-7**: Compared to [4, 7] our method improves the performance significantly with an error rate of 6.27%. To analyze the result, Figure 2 shows the weight distribution of $Conv_5$ for several epochs. In addition, Figure 3 illustrates the percentage of weights that switch between single modes at intervals of 10 epochs for different layers. Though the formation to individual modes looks fairly advanced, 29.6 % of the layer weights in $Conv_5$ switch to different clusters between epoch 110 and 130. At this point, convergence seems to be already completed, but 10.5% of the weights still change their fixed-point mode until epoch 160. While adaptation in the first and the last layer is largely done at the beginning of the training, the powerful intermediate layers keep their self-reliant weight adaptation till the end. In case of layer-independant gardient scales (as in [6, 12]), the 2-bit test error of VGG-7 is about 8%.

**DenseNet**: Our DenseNet outperforms VNQ-DenseNet and performs slitghly better than ResNet56 [5, 6] while [6] uses 448 loops of weight-partitioning and retraining to adapt the network weights. To check the flexibility of our method, we repeat the experiment with exactly the same settings and a bit size of 4. The resulting test error of 4.98% clearly outperforms the baseline of 5.68%.

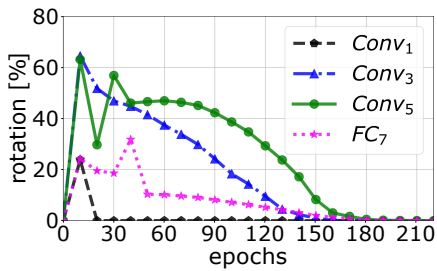

Fig. 3: Ratio of weights switching to different modes (VGG-7).

| | Model (Params) | Error | FP |
|---|---|---|---|
| BC | VGG-8 (14M) | 9.90% | ✓ |
| TWN | VGG-7 (12M) | 7.44% | ✗ |
| **SGM** | VGG-7 (12M) | **6.27%** | ✓ |
| TTQ | ResNet56 (0.85M) | 6.44% | ✗ |
| ELQ | ResNet56 (0.85M) | 6.30% | ✗ |
| VNQ | DenseNet (0.49M) | 8.83% | ✗ |
| **SGM** | DenseNet (0.49M) | **6.19%** | ✓ |

Table 2: Test error using 1-bit (BC) & 2-bit (rest) on CIFAR-10.

# 5  Conclusion

We have proposed an extension of the Gaussian prior for multiple fixed-point modes and address two common problems in weight quantization, namely high-precision scaling coefficients and incremental retraining steps. In multiple experiments, we illustrate the benefit of our layer-dependent gradient strategy and demonstrate state-of-the-art performance using ternary-valued weights.

**Acknowledgement:** We thank our colleagues Jasmin Ebert, Matthias Rath and Mark Schöne for their valuable input.